\pdfoutput=1
\documentclass[11pt]{article}

\usepackage{{ACL2023}}

\usepackage{times}
\usepackage{latexsym}

\usepackage[T1]{fontenc}
\usepackage[utf8]{inputenc}

\usepackage{microtype}

\usepackage{inconsolata}

\usepackage{multirow}
\usepackage[normalem]{ulem}
\usepackage[linesnumbered,ruled]{algorithm2e}
\usepackage{amsmath}
\usepackage{cleveref}
\usepackage{booktabs}
\usepackage{varwidth}
\usepackage{paralist}
\usepackage{enumerate}
\usepackage{amssymb}
\usepackage{pifont}
\usepackage{bm}
\usepackage{amsmath}
\usepackage{multirow}
\usepackage{caption}
\usepackage{graphicx}
\usepackage{subfigure}
\crefname{equation}{Eq.}{Eq.}
\crefname{section}{Section}{Sections}
\crefname{subsection}{Section}{Sections}
\crefname{subsubsection}{Section}{Sections}
\crefname{figure}{Figure}{Figures}
\crefname{table}{Table}{Tables}
\crefname{subfigure}{Figure}{Figures}
\crefname{algocf}{Algorithm}{Algorithms}

\newcommand{\okmark}{{\textbf{\textcolor[rgb]{0.1, 0.5, 0.1}{$\checkmark$}}}}
\newcommand{\ngmark}{{\textbf{\color{red}{\ding{55}}}}}

\makeatletter
\DeclareRobustCommand\onedot{\futurelet\@let@token\@onedot}
\def\@onedot{\ifx\@let@token.\else.\null\fi\xspace}

\def\eg{\emph{e.g}\onedot} 
\def\ie{\emph{i.e}\onedot}

\newcommand{\gptname}{GPT-4V\xspace}
\newcommand{\modelname}{MM-Navigator\xspace}

\newcommand{\red}[1]{{\color{red!50}{#1}}}
\newcommand{\blue}[1]{{\color{cyan}{#1}}}

\newcommand\blfootnote[1]{%
  \begingroup
  \renewcommand\thefootnote{}\footnote{#1}%
  \addtocounter{footnote}{-1}%
  \endgroup
}
\title{GPT-4V in Wonderland: Large Multimodal Models \\ for Zero-Shot Smartphone GUI Navigation}
\author{
  An Yan$^{\ast\diamondsuit}$,
  Zhengyuan Yang$^{\ast\spadesuit}$,
  Wanrong Zhu$^{\heartsuit}$,
  Kevin Lin$^{\spadesuit}$,
  Linjie Li$^{\spadesuit}$,
  Jianfeng Wang$^{\spadesuit}$,
  \\
  \textbf{
  Jianwei Yang$^{\spadesuit}$,
  Yiwu Zhong$^{\clubsuit}$, 
  Julian McAuley$^{\diamondsuit}$,
  Jianfeng Gao$^{\spadesuit}$,
  Zicheng Liu$^{\spadesuit}$,
  Lijuan Wang$^{\spadesuit}$}
  \\
  $^{\diamondsuit}$UC San Diego 
  $^{\spadesuit}$Microsoft Corporation
  $^{\heartsuit}$UC Santa Barbara
  $^{\clubsuit}$University of Wisconsin-Madison
  \\
  {\tt\small \{ayan,jmcauley\}@ucsd.edu, wanrongzhu@ucsb.edu, yzhong52@wisc.edu}\\
  {\tt\small \{zhengyang,keli,lindsey.li,jianfw,jianwei.yang,jfgao,zliu,lijuanw\}@microsoft.com}
}

\begin{document}
\maketitle
\blfootnote{$^\ast$ equal contributions}

\begin{abstract}
% Recent advance of Large Multimodal Models (LMMs), specifically Large Language Models equipped with visual understanding modules, has shown promising potential in various well-established benchmarks (\eg, VQA, image captioning). In this paper, we investigate the capability of \gptname for a novel task, mobile device control, and show that \gptname are decent zero-shot graphical user interface (GUI) navigators. Given an instruction, when presented with only the phone screen, \gptname is able to reason the next step action to take to complete the instruction. It also shows a strong level of visual grounding and localization of GUI screens. We start with human evaluation on a new collected iOS screen dataset, and automatic evaluation of \gptname on a recent Android screen dataset, where it outperforms previous Large Language Models with only textual inputs. We further conducted detailed analysis, shed light on future research for large vision-language models and applications to connect those models with real-world devices. 
% \gptname~\citep{openai2023gpt4,gpt4v,gpt4vcontribution,gpt4vblog,yang2023dawn}.
We present \modelname, a \gptname-based agent for the smartphone graphical user interface (GUI) navigation task. \modelname can interact with a smartphone screen as human users, and determine subsequent actions to fulfill given instructions. Our findings demonstrate that large multimodal models (LMMs), specifically \gptname, excel in zero-shot GUI navigation through its advanced screen interpretation, action reasoning, and precise action localization capabilities. We first benchmark \modelname on our collected iOS screen dataset. According to human assessments, the system exhibited a $91\%$ accuracy rate in generating reasonable action descriptions and a $75\%$ accuracy rate in executing the correct actions for single-step instructions on iOS. Additionally, we evaluate the model on a subset of an Android screen navigation dataset, where the model outperforms previous GUI navigators in a zero-shot fashion. Our benchmark and detailed analyses aim to lay a robust groundwork for future research into the GUI navigation task.
The project page is at \url{https://github.com/zzxslp/MM-Navigator}.
%We further conducted detailed analysis, shed light on future research for large vision-language models and applications to connect those models with real-world devices. 
\end{abstract}
\section{Introduction}
%% task and motivation
Building autonomous agents capable of interacting with computing devices and following human commands has been a long-standing topic in the machine learning community~\citep{bolt1980put,lieberman1995letizia}. Since the advent of smartphones, there has been a practical demand for creating virtual assistants, like Siri, Cortana, and Google Assistant, which have the potential to significantly enhance user experience and assist individuals who are physically or situationally impaired. Ideally, these assistants would competently carry out everyday tasks based on natural language instructions, ranging from simple actions like setting a timer to more complex tasks such as locating the ideal hotel for a family vacation.

%% (1,2 together, thanks to dataset work...) related works and problem; 1. vision-finetune, limited domain; 2. LLM to text, generalize but not lose information in I2T; 3. recent LMM demo on GUI Navigation, lacks quantitative evaluation => Ours; new baseline for the task
% Recent work has explored leveraging LLMs for mobile device control and task automation on smartphones following human instructions~\citep{Rawles2023AndroidIT,wen2023empowering,zhan2023you,wang2023enabling}. Since LLMs can only take text inputs, an intermediate step which converts UI information into text descriptions for LLMs to understand screens is often necessary.
Recent studies have started to explore mobile device control and smartphone task automation following human instructions~\citep{Rawles2023AndroidIT,wen2023empowering,zhan2023you,wang2023enabling}. Representative approaches include describing screen images with text and processing converted text with large language models (LLMs)~\citep{Rawles2023AndroidIT,wen2023empowering}, or training a vision-language model to generate actions in a supervised manner~\citep{Rawles2023AndroidIT,zhan2023you}. However, these supervised models, when trained on specific types of screens and instructions~\citep{Rawles2023AndroidIT}, exhibit limited effectiveness in generalizing to real-world scenarios. On the other hand, the LLM-based approaches generalize better, but the intermediate step of converting screen images to text results in information loss and consequently hurts performance.
% Since LLMs can only take text inputs, an intermediate step which converts UI information into text descriptions for LLMs to understand screens is often necessary. In this paper, we explore a different task setting with the recent advance of Large Multimodal Models (LMMs) such as \gptname, where the screen pixels are feed into the model directly, and the model is asked to predict next-step actions with the current screen and an user instruction.
Inspired by the efficacy and broad applicability of recent large multimodal models (LMMs), we explore utilizing an LMM, \gptname~\citep{openai2023gpt4,gpt4v,gpt4vcontribution,gpt4vblog,yang2023dawn}, for zero-shot smartphone GUI navigation, aiming to set a new strong baseline for this intriguing task.
% In this paper, we explore a different task setting with the recent advance of Large Multimodal Models (LMMs) such as \gptname, where the screen pixels are feed into the model directly, and the model is asked to predict next-step actions with the current screen and an user instruction.

%% Our approach; there are two required capabilities: understand screen and what to do; and convert such understanding into executable actions. For the first, text prompting; For the second, Set-of-Mark~\citep{yang2023set} prompting.
We identify two primary challenges for GUI navigation with LMMs, namely intended action description and localized action execution. First, the model should understand the screen image and text instruction input, and reason over the query to determine the appropriate action to take, such as providing a natural language description ``clicking the Amazon icon in the third row and fourth column.'' Second, the model should convert such high-level understanding into a formatted action that can be easily executed based on rules, such as ``\{\emph{Action: Click, Location: (0.31, 0.57)}\}.'' In our approach, we prompt \gptname with an image and text for action planning, and place set-of-mark tags~\citep{yang2023set} to anchor the generated outputs. Specifically, we associate these marks with spatial locations with the help of segmentation or OCR models. 
To this end, our proposed \gptname-based system, namely \modelname, can generate executable actions conditioned on the screen image, the text instruction and its interaction history.

%% Evaluation; experiment on two; iOS (our own clean, analytic dataset) is tailored for probing insights for two aspects, with the two designed sub-tasks, quantitatively reflect how reliable GPT-4V is on each aspects. Also, follow previous studies on Android benchmark for comprehensive benchmarking.
% We experiment with two real-world datasets:~(1) a small iOS dataset with screenshots and user instructions that we manually created; (2) and a large-scale Android benchmark~\citep{Rawles2023AndroidIT} which is released recently. We propose a method with Set-of-Mark~\citep{yang2023set} to efficiently leverage \gptname for this new task. Both human and automatic evaluation are conducted, and we find evidently that \gptname is a strong zero-shot GUI navigator for smartphones, significantly outperforming previous text-based LLMs.
We benchmark \modelname on two datasets. We start with an iOS GUI navigation dataset with screenshots and user instructions that we manually collected. This clean analytic dataset is designed to probe insights for the two challenges in GUI navigation: intended action description and localized action execution. Human evaluations are used to assess \gptname on these two tasks, with accuracy rates of $91\%$ and $75\%$, respectively. Additionally, we assess the model on a random subset from the recently released Android navigation benchmark~\citep{Rawles2023AndroidIT}. We follow the proposed evaluation protocol in the benchmark, together with extra human evaluations.
The strong performance demonstrates that \modelname is an effective GUI navigator for smartphones, significantly outperforming previous LLM-based approaches. We provide in-depth analyses of the representative success and failure cases. We find that the current state of \gptname may already be effective in aiding humans in various real-world GUI navigation scenarios, as evidenced by the multi-screen results in Figure~\ref{fig:ios_episode}. However, continued enhancements are still essential to further increase the system's reliability, as revealed in our analyses.
%% observation and conclusion: effective; may have failures but correct via iteration; ultimately, can already lead to the multi-screen exploration in Figure~\ref{fig:ios_episode} and video link.

Our contributions are summarized as follows.
\vspace{-4pt}
% \vspace{-3pt}
\begin{itemize}
\setlength\itemsep{-0.5pt}
    % \item We explore \gptname for smartphone GUI navigation, and present a new method with Set-of-Mark prompting. \zyang{better wrap the ``model'' with a name/story?}
    \item We present \modelname, an agent system built on \gptname for smartphone GUI navigation. \modelname effectively incorporates action histories and set-of-mark tags to produce precise executable actions.
    \item We collect a new analytic dataset with diverse iOS screens and user instructions, which evaluates two main challenges in GUI navigation with LMMs: intended action description and localized action execution.
    % \item We conduct both automatic and human evaluation on two datasets, along with the collection of a new evaluation dataset with diverse iOS screens and user instructions.
    \item We perform extensive evaluations, both automatic and human, on two datasets and provide detailed analyses. The impressive results demonstrate the effectiveness of \modelname for GUI navigation.
\end{itemize}
\section{Related Work}

% \subsection{Autonomous GUI Navigation}
\paragraph{Autonomous GUI navigation.}
Autonomous GUI navigation involves a model following instructions to maneuver through different graphical user interfaces, such as websites or applications, to perform the user-queried task. Current benchmarks collected either synthetic or real-world user-generated instructions to evaluate models' abilities in identifying specific UI elements~\citep{pmlr-v70-shi17a,li-etal-2020-mapping,Bai2021UIBertLG}, or achieving overarching task objectives by interacting with a series of GUI views~\citep{li-etal-2020-mapping,Burns2021MobileAT,Venkatesh2022UGIFUG,Deng2023Mind2WebTA,Rawles2023AndroidIT}. To understand the visual information from these GUI views, one line of work adopts a model structure that can process multimodal inputs~\citep{Sun2022METAGUITM,redmon2016you}. Other methods focus on converting the UI scene text and icons into the text-only HTML format, such as single-module LLMs can process these text inputs for GUI navigation~\citep{Zhang2021ScreenRC,Rawles2023AndroidIT,wen2023empowering}.
% attempts to detect the text with OCR and parse out UI elements, and reconstruct the text and icons to HTML syntax so as to fit the input for LLMs~\citep{Zhang2021ScreenRC,Rawles2023AndroidIT,wen2023empowering}. 

\paragraph{Multimodal agents.}
Recent advancements in LLMs~\citep{brown2020language,openai2023gpt4,chowdhery2022palm,anil2023palm,touvron2023llama,hoffmann2022training} have catalyzed the exploration of LLM-based agent systems~\citep{madaan2023self,shinn2023reflexion,pan2023automatically,yao2022react,schick2023toolformer,paranjape2023art,pryzant2023automatic,guo2023learning,zhao2023expel,yang2023large}, which integrate reasoning logic and external tools for a variety of complex language tasks. Inspired by the success in the NLP domain, multimodal researchers delve into multimodal agents. The line of research begins with LLM-based multimodal agents~\citep{gupta2023visual,suris2023vipergpt,wu2023visual,yang2023mmreact,shen2023hugginggpt,lu2023chameleon,yu2023mm,li2023multimodal}, such as MM-ReAct~\citep{yang2023mmreact} for advanced visual reasoning and Visual ChatGPT~\citep{wu2023visual} for iterative visual generation and editing. Propelled by the rapid advancements of LMMs~\citep{alayrac2022flamingo,driess2023palme,openai2023gpt4,gpt4v,gpt4vcontribution,gpt4vblog,yang2023dawn,bard}, the latest studies have begun to investigate the LMM-powered multimodal agents~\citep{yang2023idea2img,liu2023llavaplus}, thereby surpassing the need for basic visual description tools like caption models~\citep{wang2022git,wu2022grit}. Our proposed methodology represents a specialized LMM-based agent for GUI navigation. We aim to provide a comprehensive analysis and a strong baseline for this task.
% \section{Method}
\section{\modelname}
\subsection{Problem Formulation}
When presented with a user instruction $X_{instr}$ in natural language, the agent is asked to complete a series of actions on the smartphone to complete this instruction. The entire process of agent-environment interactions from initial to final states is called an episode. At each time step $t$ of an episode, the agent will be given a screenshot $I^t$, and decide the next step action to take in order to complete the task.
%in the end. 
% We identify two main challenges:~(1) 

% \subsection{Interacting with \gptname}
% Prompt Engineering has been an emergent topic for natural language processing and large language models~\citep{wei2022chain,kojima2022large}. With the development of recent instruction-following models~\citep{openai2023gpt4,touvron2023llama}, language models have been less dependant on the prompts. In this paper, we are mainly interested in exploring how to use \gptname for smartphone navigation and designing methods to enable this function, and do not focus on finding optimal prompt designs. Nevertheless, we performed ablation studies and robustness check on different prompt templates we used, as shown in~\cref{sec:android_ablation}.

\subsection{Screen Grounding and Navigation via Set of Mark}
\gptname serves as a multimodal model that takes visual images and text as inputs and produces text output. One challenge is how do we communicate with \gptname to perform actions on screen. A possible solution is to ask the model to reason about coordinates to click given a screen. However, based on our preliminary exploration, though \gptname have a good understanding of the screen and approximately where to click to perform an instruction by describing the corresponding icon or text, it appears to be bad at estimating accurate numerical coordinates.  

Therefore, in this paper, we seek a new approach, to communicate with \gptname via Set-of-Mark prompting~\citep{yang2023set} on the screen. Specifically, given a screen, we will detect UI elements via the OCR tool and IconNet~\citep{sunkara2022towards}. Each element has a
bounding box and either OCR-detected text or an icon class label (one of the possible 96 icon types detected by~\citep{sunkara2022towards}) are contained. At each step time $t$, we add numeric tags to those elements, and present \gptname with the original screen $I^t$ and the screen with tags $I^t_{tags}$. The output text $Y_{action}$ of \gptname will be conditioned on the two images. If \gptname decides to click somewhere on the screen, it will choose from the available numeric tags. In practice, we found this simple method works well, setting up a strong baseline for screen navigation with large multimodal models. 
% An illustration of the method is shown in.

\subsection{History Generation via Multimodal Self Summarization}
Set-of-Mark prompting bridges the gap between text outputs from \gptname and executable localized actions. However, the agent's ability to maintain a historical context is equally important in successfully completing tasks on smartphones. 
% the agent should also be able to keep track of its history. 
% The challenge rises where we need to design a method such that for each time step of an episode, the agent can efficiently reason the next step action based on its history and current screen. 
The key difficulty lies in devising a strategy that allows the agent to effectively determine the subsequent action at each stage of an episode, taking into account both its prior interactions with the environment and the present state of the screen.
The naive approach of feeding all historical screens or actions into the agent is computationally expensive and may decrease the performance due to information overload. For example, screens at each step can change rapidly, and most of the historical screen information is not useful for reasoning about future actions. 
Humans, on the other hand, can keep track of a short memory of the key information after performing a sequence of actions. We aim to find a more concise representation than a sequence of screens or actions. Specifically, at each time step, we ask \gptname to perform multimodal self summarization, which converts the historical actions and current step information into a concise history in the form of natural language, which is formulated as follows:
\begin{align}
    Y^t_{action} = \Theta_{gpt}(X_{instr}, I^t, I^t_{tags}, Y^{t-1}_{history}) \\
    Y^t_{history} = \Theta_{gpt}(Y^{t}_{action}, Y^{t-1}_{history})
\end{align}
where $Y^t_{action}$ is the action to take at current step $t$, $Y^{t}_{history}$ is the summarized history based on $Y^t_{action}$ and $Y^{t-1}_{history}$, $\Theta_{gpt}$ is the parameterized \gptname model. In this way, the trace of history will be generated auto-regressively when an episode is rolled out.

%%%%%%%
%% iOS dataset analysis
\section{iOS Screen Navigation Experiment}
\begin{table}[t]
\centering
{\small
% \vspace{-6mm}
\caption{Zero-shot \gptname iOS screen navigation accuracy on the ``intended action description'' and ``localized action execution'' tasks, respectively.\label{tab:iosexp}}
% \vspace{-6mm}
\setlength{\tabcolsep}{3pt}
    \begin{tabular}{lcc}
    \toprule
      Setting & \multicolumn{2}{c}{Accuracy}\\
    \midrule
     Intended Action Description & 50/55\ \ = & 90.9\%  \\
     Localized Action Execution &  41/55\ \ = & 74.5\%  \\
    \bottomrule
  \end{tabular}
  }
  % \vspace{-6mm}
\end{table}
%%%%%%%%%%%%%%%%%%%%%%%%%%%%%%%%%%%%%%%%
\begin{figure*}[t]
% \vspace{0.1in}
\centering
\centerline{\includegraphics[width=1.2\textwidth]{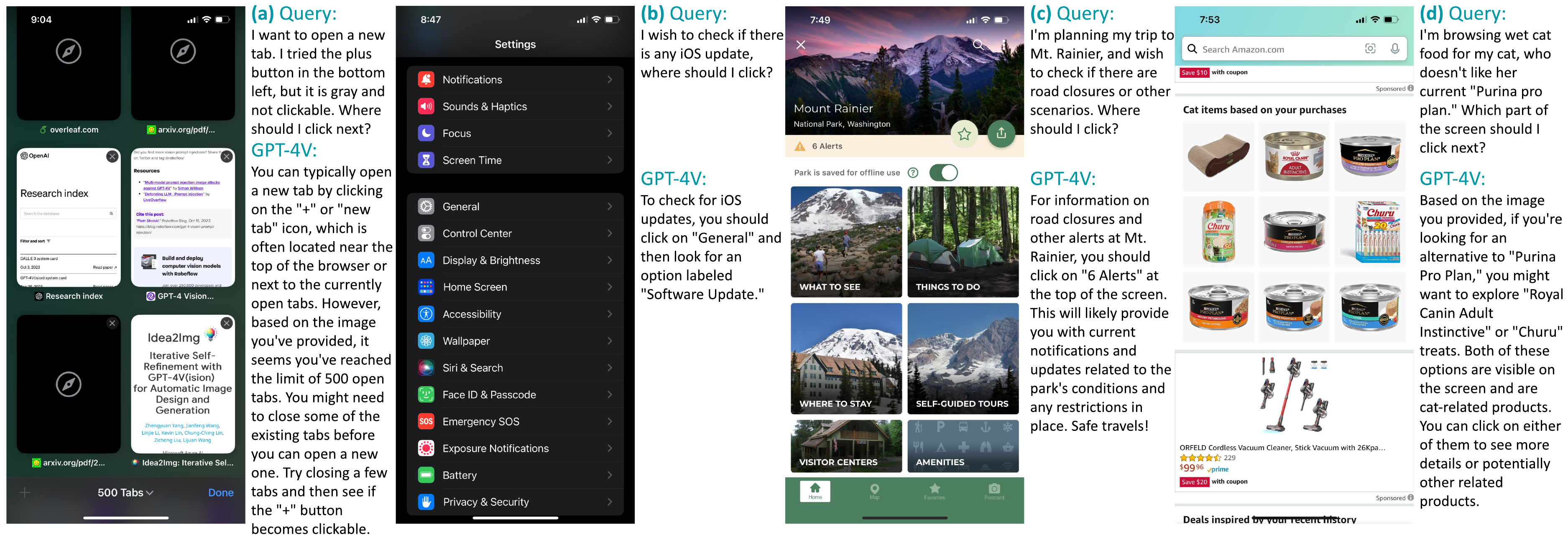}}
% \vspace{0.1in}
\caption[Caption for LOF]{
    Intended action description examples. Best viewed by zooming in on the screen.
	}
 % \vspace{0.1in}
\label{fig:ios_understand}
\end{figure*}
%%%%%%%%%%%%%%%%%%%%%%%%%%%%%%%%%%%%%%%%
\subsection{Experimental Setup}
\paragraph{Dataset}
We begin by conducting analytical experiments on iOS screens to understand \gptname's capability in GUI navigation. Successfully operating smartphones in a human-like manner involves different types of screen understanding abilities. Firstly, there is the semantic reasoning ability, which involves comprehending screen inputs and articulating the necessary actions to fulfill given instructions. Secondly, there is the need to translate these action descriptions into specific localized actions, such as determining the precise location for a screen click. Correspondingly, we develop two sets of test screens to disentangle these two aspects, which are referred to as ``intended action description'' and ``localized action execution,'' respectively.
% understanding screen, and perform localized actions. To perform GUI navigation, requires semantic understanding and localized action predictions. Two separate sets to disentangle these two aspects.

In this study, we gather $110$ instructions from $6$ human annotators, evenly divided into two distinct sets, containing iOS screens with and without added marks. The first set, ``intended action description,'' involves \gptname taking an iOS screenshot image and an instruction as inputs, and generating an open-ended text description of the desired action to perform. This set aims to assess \gptname's ability to reason the correct action to perform. Moving beyond having someone click the screen for \gptname~\citep{yang2023dawn,lin2023mm}, we investigate directly generating formatted executable actions. In the second set, ``localized action execution,'' we add marks~\citep{yang2023set} to ground screen locations with interactive SAM~\citep{kirillov2023segment}, and let \gptname use the mark indexes to perform localized actions. Other approaches, such as textualized box coordinates~\citep{chen2021pix2seq,yang2022unitab,wang2022ofa}, screen visual grounding~\citep{yu2016modeling,mao2016generation,plummer2015flickr30k,yang2019fast,deng2021transvg}, object detectors~\citep{ren2015faster,carion2020end} could also translate action descriptions into executable actions. % We postpone these discussions to Section~\ref{sec:grounding}.

\paragraph{Human evaluation metrics.}
We use human evaluation for the analytical experiments on iOS screens, with a binary score for each sample indicating if the output is correct. For ``intended action description'', human annotators determine if the output text description could lead to the correct output.
% as if humans help \gptname perform its described action. 
For ``localized action execution,'' human annotators assess if clicking the location (\ie, location of the selected mark) fulfills the given instruction. Each sample is assigned a binary score, either 0 or 1, to reflect its correctness.

%%%%%%%%%%%%%%%%%%%%%%%%%%%%%%%%%%%%%%%%
% \subsection{Visual Localization and Action}
\begin{figure*}[t]
% \vspace{0.1in}
\centering
\centerline{\includegraphics[width=1.2\textwidth]{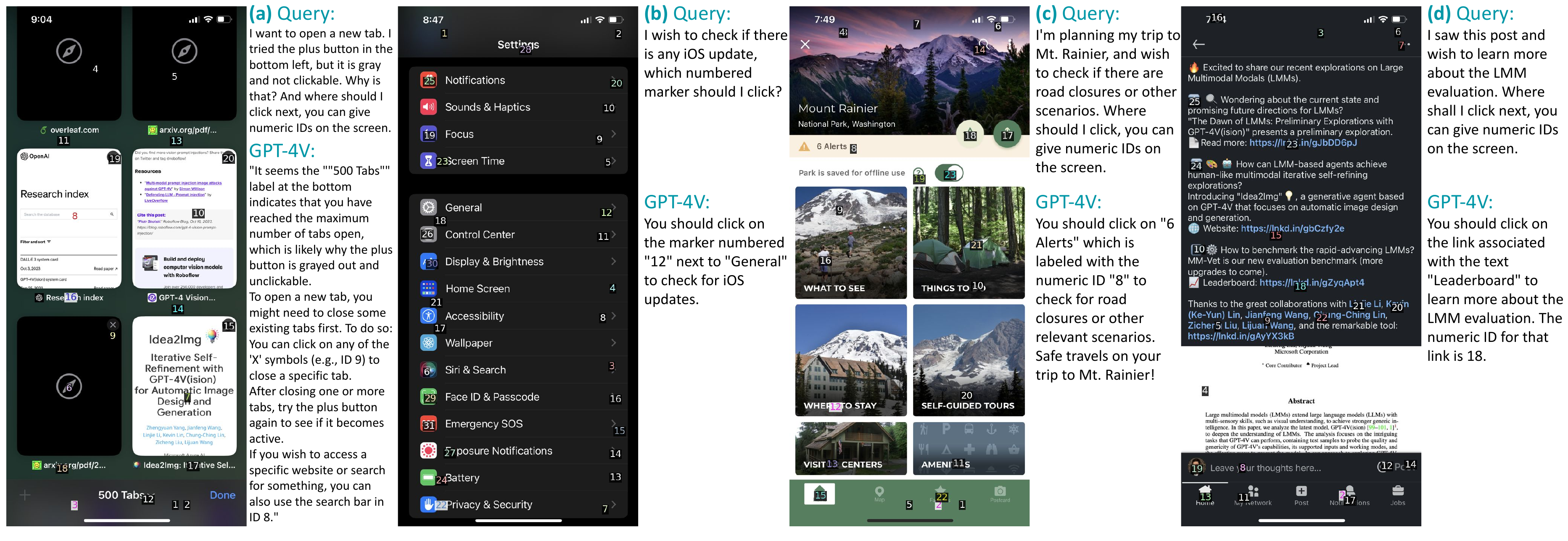}}
% \vspace{0.1in}
\caption[Caption for LOF]{
    Localized action execution examples. Best viewed by zooming in on the screen.
	}
 % \vspace{0.1in}
\label{fig:ios_localization}
\end{figure*}
%%%%%%%%%%%%%%%%%%%%%%%%%%%%%%%%%%%%%%%%
\begin{figure*}[t]
% \vspace{0.1in}
\centering
\centerline{\includegraphics[width=1.2\textwidth]{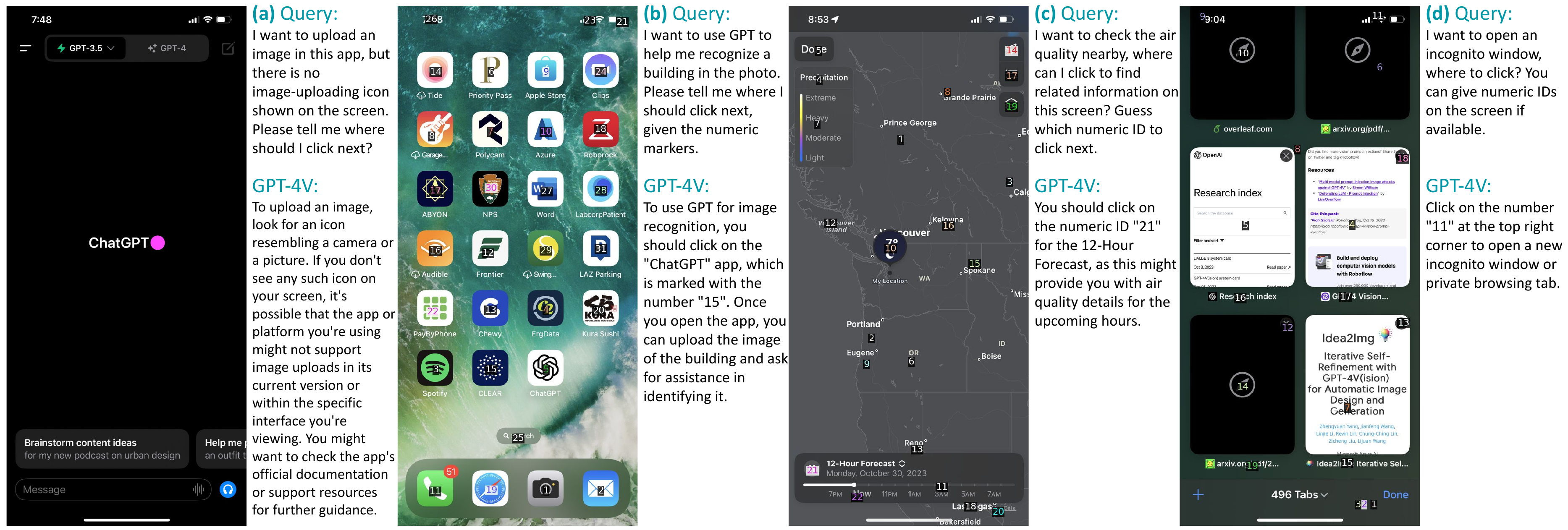}}
% \vspace{0.1in}
\caption[Caption for LOF]{
    Representative failure cases in iOS screen navigation. Best viewed by zooming in on the screen.
	}
 % \vspace{0.1in}
\label{fig:ios_failure}
\end{figure*}
%%%%%%%%%%%%%%%%%%%%%%%%%%%%%%%%%%%%%%%%
\subsection{Intended Action Description}
%%%%%%%%%%%%%%%%%%%%%%%%%%%%%%%%%%%%%%%%
Table~\ref{tab:iosexp} reports an accuracy of $90.9\%$ on generating the correct intended action description, quantitatively supporting \gptname's capability in understanding the screen actions to perform~\citep{yang2023dawn,lin2023mm}.
Figure~\ref{fig:ios_understand} showcases representative screen understanding examples. Given a screen and a text instruction, \gptname~gives a text description of its intended next move. For example, in Figure~\ref{fig:ios_understand}(a), \gptname understands the Safari browser limits of ``the limit of 500 open tabs,'' and suggests ``Try closing a few tabs and then see if the "+" button becomes clickable.'' Another example is telling the procedure for iOS update: ``You should click on "General" and then look for an option labeled "Software Update'' in (b). \gptname also effectively understands complicated screens with multiple images and icons. For example, in (c), \gptname mentions, ``For information on road closures and other alerts at Mt. Rainier, you should click on "6 Alerts" at the top of the screen.'' Figure~\ref{fig:ios_understand}(d) gives an example in online shopping, where \gptname suggests the correct product to check based on the user input of the desired ``wet cat food.''
% GUI understanding, infer road condition might be in alerts. Shopping: after product brands and recommend the one to view.

\subsection{Localized Action Execution}
%%%%%%%%%%%%%%%%%%%%%%%%%%%%%%%%%%%%%%%%
A natural question is how reliable \gptname can convert its understanding of the screen into executable actions. Table~\ref{tab:iosexp} shows an accuracy of $74.5\%$ on selecting the location that could lead to the desired outcome. 
Figure~\ref{fig:ios_localization} shows the added marks with interactive SAM~\citep{yang2023set,kirillov2023segment}, and the corresponding \gptname outputs. As shown in Figure~\ref{fig:ios_localization}(a), \gptname can select the ``X'' symbol (ID: 9) to close the tabs, echoing its previous description in Figure~\ref{fig:ios_understand}(a). \gptname is also capable of selecting the correct location to click from the large portion of clickable icons, such as the screen shown in (b). Figure~\ref{fig:ios_understand}(c) represents a complicated screen with various images and icons, where \gptname can select the correct mark $8$ for the reading the ``6 Alerts.'' Within a screen with various texts, such as (d), \gptname can identify the clickable web links, and locate the queried one with the correct position $18$.
% Localize the icon on a complex screen with lots of texts, with the target link among them.

\begin{figure*}[t]
\vspace{-0.1in}
\centering
% \centerline{\includegraphics[width=1.2\textwidth]{figures/ios/demo.pdf}}
\centerline{\includegraphics[width=0.85\textwidth]{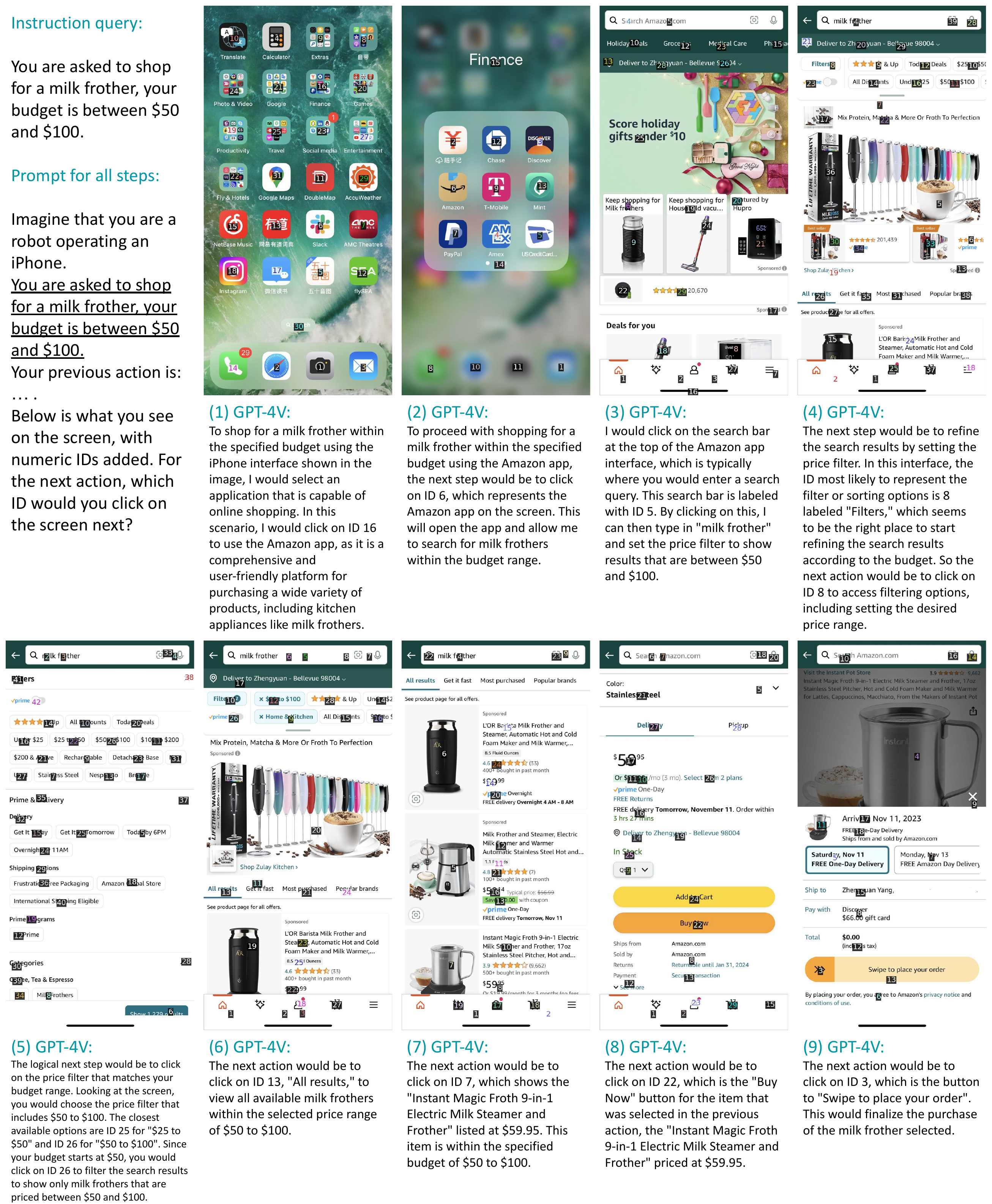}}
% \vspace{0.1in}
\caption[Caption for LOF]{
    Episode examples on iOS screen navigation. Best viewed by zooming in on the screen.% Video link:.
	}
 \vspace{-0.1in}
\label{fig:ios_episode}
\end{figure*}

%%%%%%%%%%%%%%%%%%%%%%%%%%%%%%%%%%%%%%%%
% \subsection{Quantitative Performance}
\subsection{The Current State with \gptname}
% \paragraph{Quantitative performance.} A table of (xx/60) here.
From the analytical experiments on iOS screens, we find \gptname is capable of performing GUI navigation. Although several types of failure cases still occur, as outlined below, \modelname shows promise for executing multi-screen navigation to fulfill real-world smartphone use cases. We conclude the section with qualitative results on such episode-level navigation queries.
% However, there still exist various kinds of failure cases, as we detail below. Nonetheless, \modelname may already be helpful in performing multi-screen navigation to fulfill real-world cellphone use cases. We conclude the section with episode-level navigation results.%qualitative examples of \gptname performing multi-screen navigation to fulfill real-world cellphone use cases.

\paragraph{Failure cases.}
%%%%%%%%%%%%%%%%%%%%%%%%%%%%%%%%%%%%%%%%
Despite the promising results, \gptname does make errors in the zero-shot screen navigation task, as shown in Table~\ref{tab:iosexp}. These errors are illustrated through representative failure cases as follows. \textbf{(a)} \gptname might not generate the correct answer in a single step when the query involves knowledge the model lacks. For example, \gptname is not aware that only ``GPT-4'' can support image uploads, hence it fails to click the ``GPT-4'' icon before attempting to find the image uploading function. \textbf{(b)} Although usually reliable, \gptname might still select the incorrect location. An example of this is selecting the mark $15$ for the ``ChatGPT'' app instead of the correct mark $5$. \textbf{(c)} In complex scenarios, \gptname's initial guess might not be correct, such as clicking the ``numeric ID 21 for the 12-Hour Forecast'' instead of the correct answer of mark 19. \textbf{(d)} When the correct clickable area is not marked, like a ``+'' icon without any marks, \gptname cannot identify the correct location and may reference an incorrect mark instead. Finally, we note that many of those single-step failures may be corrected with iterative explorations, leading to the correct episode-level outcome.
% Finally, the correct location to click may not be associated with one of the placed marks, \eg, the ``+'' icon with no marks. In this case, \gptname can not select the correct location and refers to a wrong mark instead. % interface has an oracle, that the position to click is not marked, though \gptname may sometimes be able to reflect that.

% Some complicated situations might need further exploration to obtain the knowledge that only \gptname supports image uploading at this time point. Might still fail, even for a standard home screen, although not frequent as shown in table. For challenging cases, the first trial might not be correct, as human users, as we will discuss in the next section of Android screens (the correct answer is marker 19). The marker interface has an oracle, that the position to click is not marked, though \gptname may sometimes be able to reflect that. 

% \paragraph{Complete episode example.}
\paragraph{From single screens to complete episodes.}
% Later convert to GIF.
% \zyang{also/or highlight elsewhere, e.g., teaser}
% Overall, reliable enough, single-step failures may be corrected by iterative explorations.
% Overall, 
\modelname shows an impressive capability in performing GUI navigation in a zero-shot manner. We further extend \modelname from processing a single cellphone screen to recursively processing an episode of screen inputs. Figure~\ref{fig:ios_episode} shows the qualitative result. In each step, we include the objective, ``You are asked to shop for a milk frother, your budget is between \$50 and \$100.'' and its previous action in the prompt to \gptname. We show that the model can effectively perform multi-step reasoning to accomplish the given shopping instruction. 
% For future works, it is promising to have phones automatically take the predicted actions and automate the process, thereby facilitating real-world applications and enabling episode-level evaluations.

%%%%%%%
%% Android dataset analysis
\section{Android Screen Navigation Experiment}
\subsection{Experimental Setup}
\paragraph{Dataset.}
We use the AITW dataset~\citep{Rawles2023AndroidIT} for our evaluation on Android screen navigation. AITW is a large-scale
benchmark dataset for UI control, which contains natural language instructions, screenshots on different Android systems with different resolutions, and user-annotated actions. It covers diverse multi-step tasks such as various web and application operations, app installation, and tasks with Google apps, with 715K episodes and 30K unique instructions in total. 
Table~\ref{tab:data_stats} shows the basic statistics of the dataset. We follow the split from previous work~\citep{zhan2023you}. 
% Due to the cost of calling \gptname, to this point, we used 300 episodes from the test split as our test set, which is of similar size as previous work~\citep{Rawles2023AndroidIT} where they evaluated 288 episodes with PaLM 2. 
Following the previous experiment setting~\citep{Rawles2023AndroidIT} that evaluates PaLM 2 on a randomly sampled 288 episodes, we sample 300 episodes from the test split as our test set.
\begin{table}[t]
\centering
{\small
% \vspace{-6mm}
\caption{The Android in the Wild (AITW) dataset~\citep{Rawles2023AndroidIT} statistics\label{tab:data_stats}.}
% \vspace{-6mm}
 \setlength{\tabcolsep}{2pt}
    \centering
    \begin{tabular}{lrrr}
    \toprule
     {Dataset}  & Episodes & Screens  &  Instructions\\
    \midrule
     General & 9,476 & 85,413 & 545 \\
     Install &  25,760 & 250,058 & 688 \\
     GoogleApps &  625,542 & 4,903,601 & 306  \\
     Single & 26,303 & 85,668 & 15,366 \\
     WebShopping & 28,061 & 365,253 & 13,473 \\
    \bottomrule
  \end{tabular}
  }
  \vspace{-3mm}
\end{table}
\begin{table*}[t]
\centering
\caption{Main results (\%). Segment 1: fine-tuned Llama 2 baseline; Segment 2: in-context learning LLM baselines. ``ZS'' stands for ``zero-shot.''; Segment 3: GPT-4V zero-shot results: ``image-only'' means only screen images are fed into the agent. ``+text'' adds parsed screen descriptions. ``+history'' allows the agent to access its history actions.  ``Training Free'' means a model with zero-shot performance or in-context learning. ``Text Free'' means no parsed screen description is needed. The overall score is computed as the average over all the subsets.
}
\small % 9 9
% \fontsize{8.3pt}{\baselineskip}\selectfont % font size
\renewcommand\tabcolsep{3pt} % column space

\begin{tabular}{l|cc|c|ccccc} 
\toprule
 Model & Training Free & Text Free & Overall & General & Install & GoogleApps & Single & WebShopping \\
 \midrule
Fine-tuned Llama 2 & \ngmark & \ngmark & 28.40 & 28.56 & 35.18 & 30.99 & 27.35 & 19.92\\ 
\midrule 
PaLM 2 ZS & \okmark & \ngmark& 30.90	&  -  &  - &  - &  -  & - \\
PaLM 2 5-shot & \okmark & \ngmark& 39.60	&  -  &  - &  - &  -  & - \\
ChatGPT 5-shot & \okmark& \ngmark & 7.72 &	5.93	&4.38	&10.47	& 9.39	& 8.42\\
\midrule
\gptname ZS image-only & \okmark & \okmark & 50.54 & 41.66 & 42.64 & \textbf{49.82} & 72.83  & 45.73 \\
\gptname ZS +text & \okmark & \ngmark & 51.92 & 42.44 & \textbf{49.18} & 48.26 & 76.34 & 43.35 \\
\gptname ZS +history & \okmark & \ngmark & \textbf{52.96} & \textbf{43.01} & 46.14 & 49.18 & \textbf{78.29} & \textbf{48.18} \\
\bottomrule
\end{tabular}
 \label{tab:main_results}
\end{table*}

\paragraph{Metrics.}
Following previous work~\citep{Rawles2023AndroidIT,zhan2023you}, we compute the screen-wise partial action matching score as the main evaluation metric, defined as the number of correct actions divided by the episode length, then this score is averaged over all tested episodes. A predicted action from \gptname is considered correct if both the action type and gesture match the gold ones, i.e., user actions. For click actions, it is considered correct if the selected element falls within a 14\% screen distance from the gold gestures or occurs within the same detected bounding box with user gestures. For scroll actions, it is considered correct if the selected direction has the same scroll direction (up, down, left, and right) as user gestures. The partial score has been shown to correlate with the task complete score
estimated by human evaluations~\citep{Rawles2023AndroidIT} to measure the action success rate of this task.

\paragraph{Baselines.}
We compare with the following baselines~\citep{Rawles2023AndroidIT,zhan2023you}:% with recent LLMs:
\vspace{-4pt}
\begin{itemize}
\setlength\itemsep{-1pt}
    \item PaLM-2 ZS~\citep{Rawles2023AndroidIT}: Zero-shot performance with PaLM-2~\citep{anil2023palm}, by feeding a textual description of the screen and ask it to predict an action among the supported actions in AITW. We adopt a previously proposed LLM-based design for device control~\citep{wang2023enabling}, where the input screen description is converted to HTML syntax.
    \item PaLM-2 5-shot~\citep{Rawles2023AndroidIT}: Five examples of navigation are designed as Chain-of-thought prompts. The history of prior actions taken by the agent is also fed into the model input. 
    \item ChatGPT 5-shot~\citep{zhan2023you}. The input prompts are of the same format as PaLM-2 5-shot. Experiments are conducted via the ChatGPT API.
    \item Fine-tuned Llama-2~\citep{zhan2023you}: Fine-tuning Llama-2 model~\citep{touvron2023llama} with LoRA~\citep{hu2021lora}, by feeding the model with the user instruction and screen descriptions in HTML syntax (the same that are used for in-context learning LLMs) and predict user actions. The model is fine-tuned with $1\%$ randomly sampled training data to help adapt to this task.
    
\end{itemize}

\subsection{Performance Comparison}
Our main results are shown in~\cref{tab:main_results}. % \gptname outperforms previous LLMs that take ground-truth descriptions of the screens as inputs. 
First, \gptname outperforms previous LLMs that take ground-truth descriptions of the screens as inputs. Compared with previous text-only LLMs, taking screen images as visual inputs provides an easier way for human-model interactions. It also better preserves the screen information and avoids the information loss when converting screens to text descriptions. 
Additionally, adding screen descriptions still improves the performance of \gptname. Giving the agent access to its historical interactions is helpful for better conditioned and grounded generation, and our in-context self-summarization module provides an efficient way to achieve this. 
Overall, we find \gptname presents a strong level of screen understanding of icons and text, showing the potential of visual-based device control with LMMs. 
\begin{table}[t]
\centering
\caption{Ablation studies on different tagging methods.
}
\small % 9 9
% \normalsize % 10 10
% \fontsize{8.3pt}{\baselineskip}\selectfont % font size
\renewcommand\tabcolsep{2.5pt} % column space
\begin{tabular}{l|c|ccccc} 
\toprule
 Model & Overall & General & Install & Apps & Single & Webshop \\
 \midrule
By side & 48.39 & 35.24 & 42.18 & 42.46 & 81.50 & 40.53\\ 
\midrule
Red & 49.05 & 41.61 & 35.00 & 43.81  & 76.50 & 48.32 \\
\midrule 
Center	&  49.72  & 47.93 & 36.06 & 44.54 & 79.50  & 40.58 \\
\bottomrule
\end{tabular}
 \label{tab:ablation_tags}
\end{table}

\begin{table}[t]
\centering
\caption{Ablation studies on different prompts.
}
\small % 9 9
% \normalsize % 10 10
% \fontsize{8.3pt}{\baselineskip}\selectfont % font size
\renewcommand\tabcolsep{2.5pt} % column space
\begin{tabular}{l|c|ccccc} 
\toprule
 Model & Overall & General & Install & Apps & Single & Webshop \\
 \midrule
Baseline& 48.39 & 35.24 & 42.18 & 42.46 & 81.50 & 40.53\\ 
\midrule 
Think	&  46.66  & 35.01 & 39.12 & 40.50  & 76.50 & 42.18 \\
\midrule
Specific & 48.77 & 50.77 & 40.54 & 42.32 & 69.50  & 40.71 \\
\bottomrule
\end{tabular}
 \label{tab:ablation_prompts}
\end{table}
\subsection{Ablation Studies}
\label{sec:android_ablation}
For the ablation studies, we randomly sampled 50 episodes in total from 5 categories, which is a different subset used by the main results. 
\paragraph{Different tagging methods.}
We first perform an ablation study to compare the performance with different methods to add tags on screen, shown in~\cref{tab:ablation_tags}. We consider three methods:~(1) \textit{By side} which adds tags with black squares (same style as~\citep{Rawles2023AndroidIT} by the left side of each detected icon;~(2) \textit{Red} which uses red circles for each tag;~(3) \textit{Center} which adds tags with black squares at the center of each detected box. First, adding tags by the left side of boxes may cause problems, for example, some icons may be too close to each other, hence leading to slightly worse results. For tagging styles, we didn't find a significant difference between red cycles and black rectangles, though empirically black rectangles~\citep{yang2023set} perform slightly better.

\paragraph{Different prompts.}
We then perform robustness check with different prompting variants:~(1) \textit{Baseline}: Simply ask \gptname to take actions;~(2) \textit{Think}: Prompt \gptname to think step by step~\citep{kojima2022large};~(3) \textit{Detail}: Provide more context for this task. Overall, we did not observe improvements by ``thinking step by step'', but adding more task descriptions helps \gptname to better execute actions.
% The prompt templates are in~\cref{sec:appendix}

\subsection{Error Analysis}
We look into \gptname prediction traces and attempt to categorize common types of errors that cause mismatching between \gptname predictions and human annotations.

\begin{figure}[]
\centering
\centerline{\includegraphics[width=\linewidth]{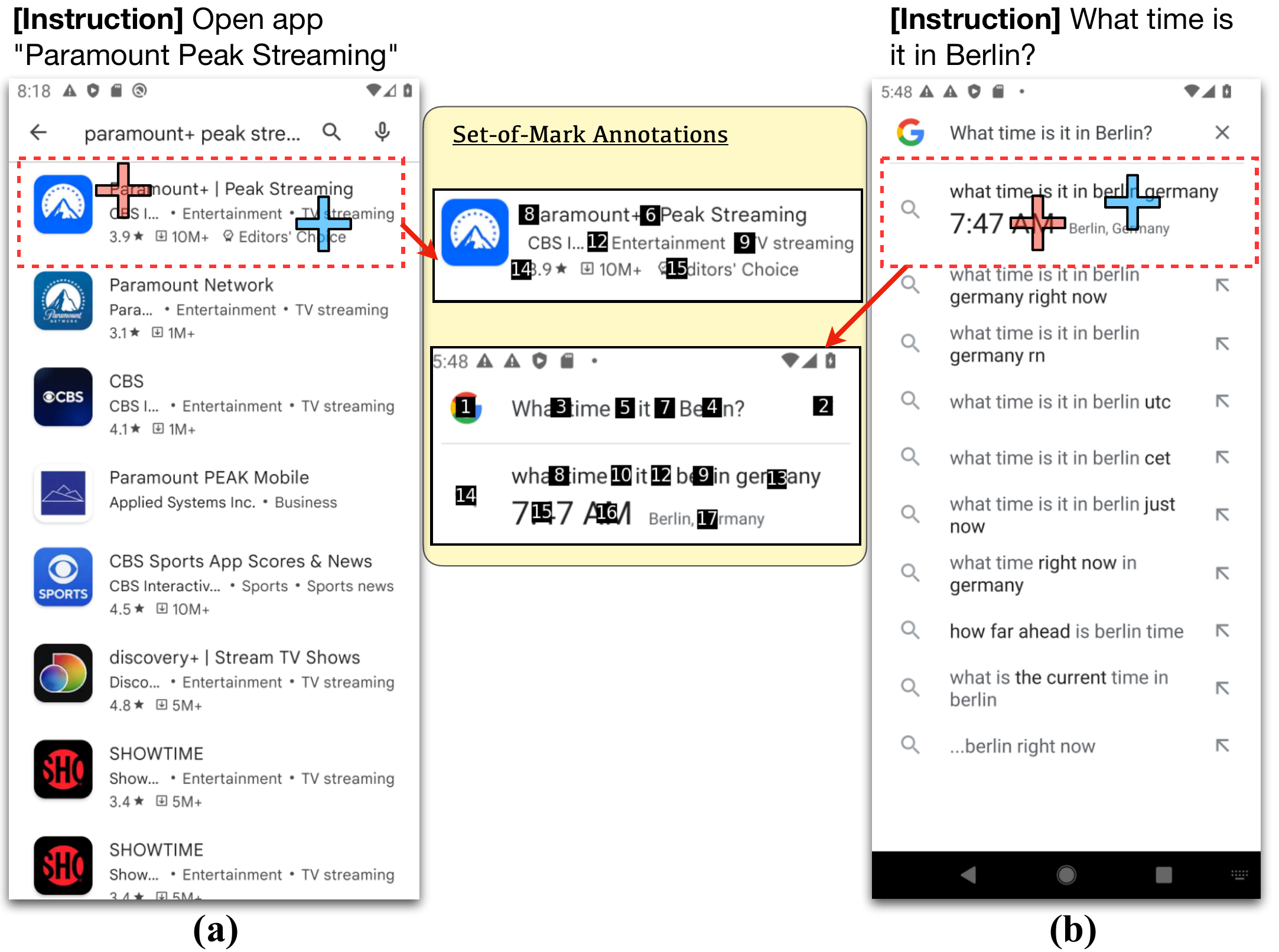}}
\caption[Caption for LOF]{
    Examples of false negatives that are caused by inaccurate parsing in Set-of-Mark annotations. ``\blue{+}'' denotes human annotation, and ``\red{+}'' is \gptname prediction. 
	}
\label{fig:error_false_negative_1}
\vspace{-0.1in}
\end{figure}

\begin{figure}[]
\centering
\centerline{\includegraphics[width=\linewidth]{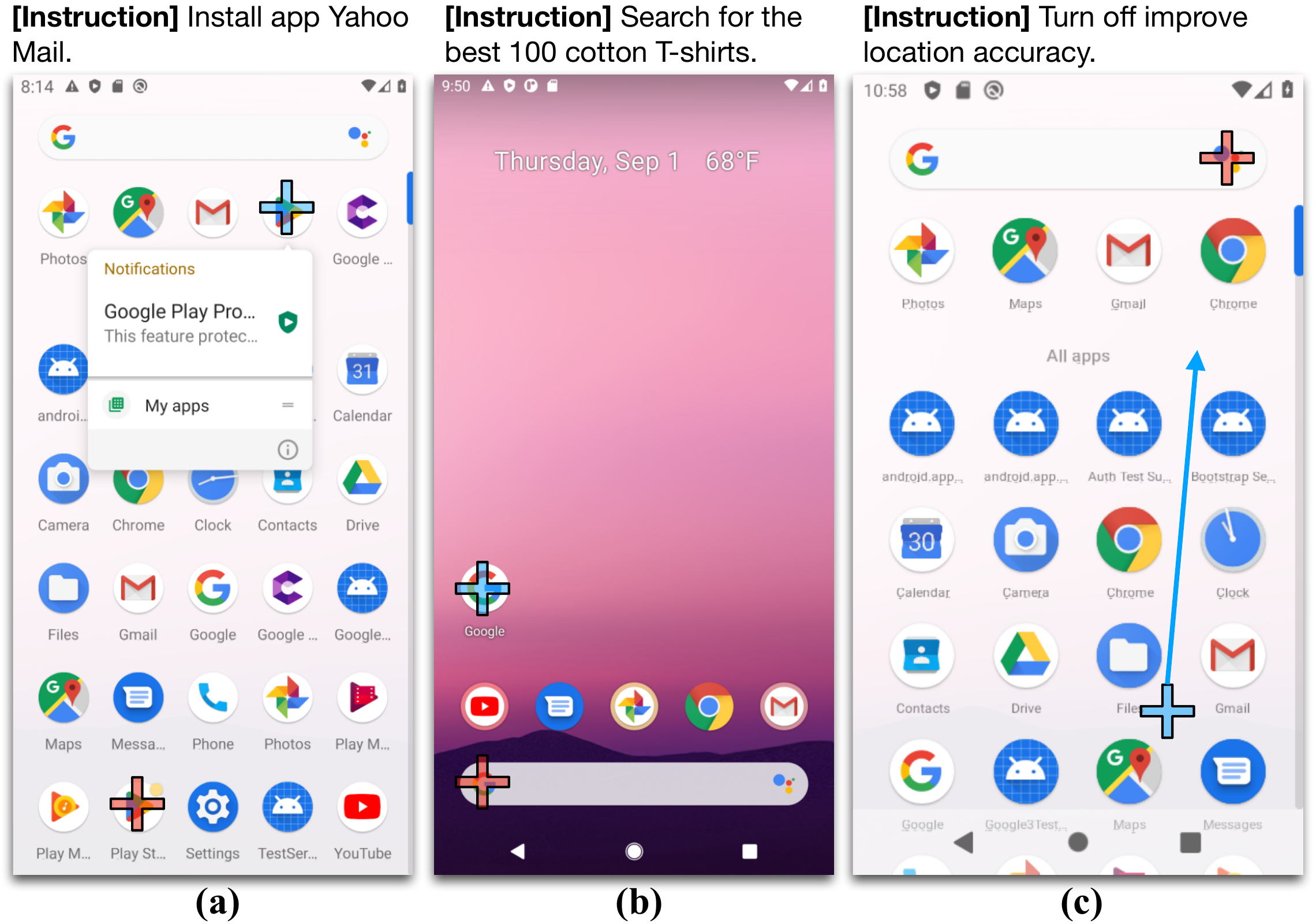}}
\caption[Caption for LOF]{
    Examples of false negative scenarios that are caused by imperfections in ground truth dataset annotations.
    ``\blue{+}'' denotes human annotation, ``\blue{$\nearrow$}'' shows the trace of scrolling, and ``\red{+}'' is \gptname prediction. 
	}
\label{fig:error_false_negative_2}
\vspace{-0.1in}
\end{figure}

We notice false negative cases where the mismatches are rooted in inaccurate Set-of-Mark~\citep{yang2023set} annotation parsing or imperfect dataset annotation. In these cases, the predictions made by \gptname are correct after manual justification, but are classified as wrong predictions in automatic evaluation because the target regions are over-segmented (\eg, Figure~\ref{fig:error_false_negative_1}(a)(b)), or because the ground-truth annotation only covers one of the many valid actions (\eg, Figure~\ref{fig:error_false_negative_2}(a) has two Google Play logo; Figure~\ref{fig:error_false_negative_2}(b) has multiple ways of accessing Google Search; and users may lookup ``Setting'' by direct search as \gptname, or by scrolling down as the human annotation in Figure~\ref{fig:error_false_negative_2}(c)).

\begin{figure}[]
\centering
\centerline{\includegraphics[width=\linewidth]{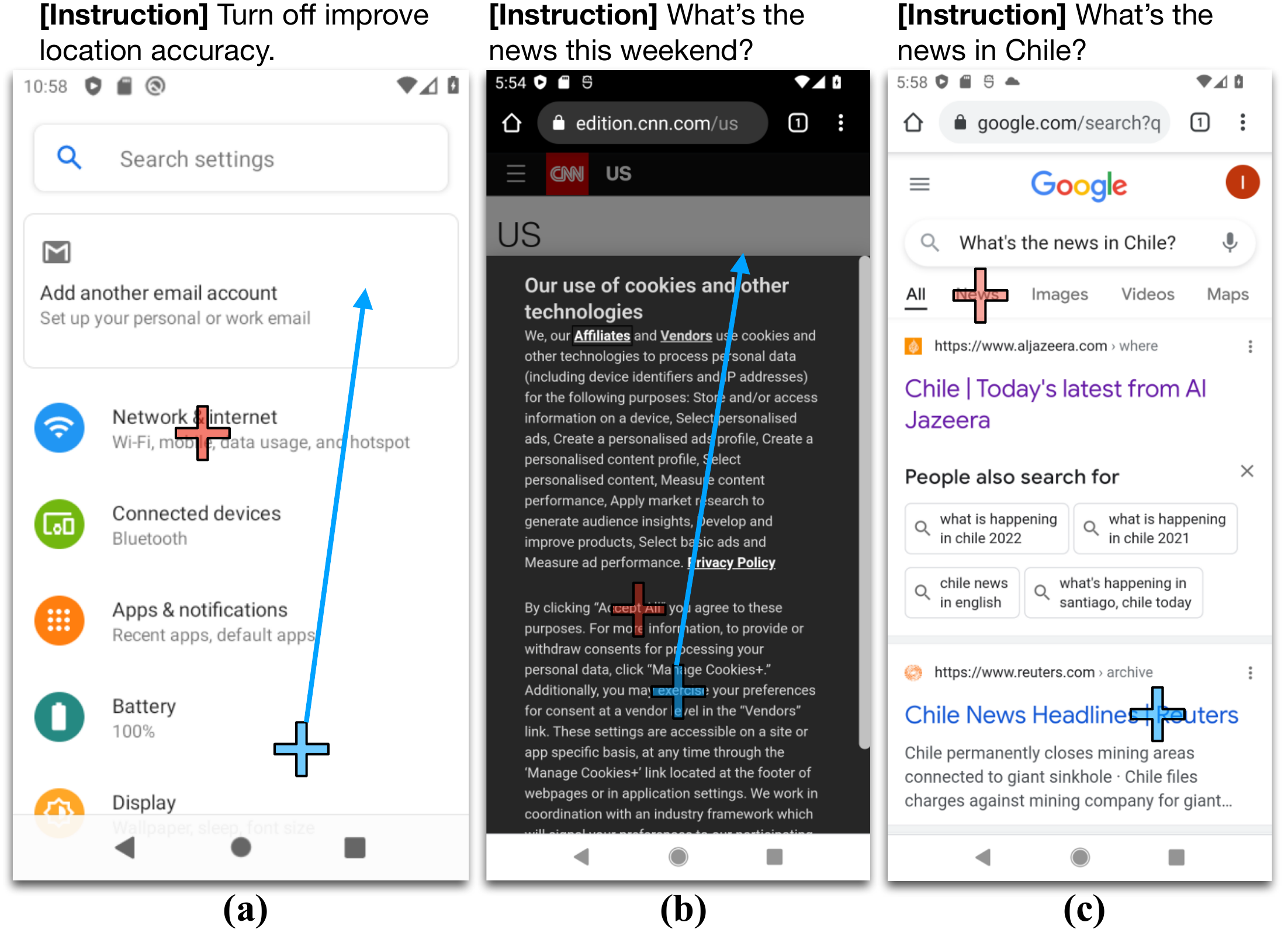}}
\caption[Caption for LOF]{
    Examples of true negative cases where \gptname makes mistakes.
    ``\blue{+}'' denotes human annotation, ``\blue{$\nearrow$}'' shows the trace of scrolling, and ``\red{+}'' is \gptname prediction. 
	}
\label{fig:error_true_negative}
\vspace{-0.2in}
\end{figure}

Figure~\ref{fig:error_true_negative} shows a few true negative examples of \gptname failing the designated tasks. In our zero-shot testing setup, \gptname is not provided with demonstrative examples to learn user action patterns. In this case, while users may scroll down or up to explore the GUI, we notice \gptname is more likely to perform the action of ``click'' on each screen, leading it to occasionally make short-sighted decisions. In Figure~\ref{fig:error_true_negative}(a), \gptname attempts to look for ``improve location accuracy'' in ``Network\&Internet'' among the listed visible tabs, while the user decides to scroll down and look for more aligned setting tabs.  In Figure~\ref{fig:error_true_negative}(b), \gptname clicks on ``Accept All'', which is not a button. In Figure~\ref{fig:error_true_negative}(c), \gptname also shows a more literal understanding of the instruction and the current observation as in (b), clicking the ``News'' tab in the Google Search platform instead of actually visiting the news website. 

\section{Discussion}
\paragraph{Future benchmarks for device-control.}
For future benchmarks, more dynamic interaction environments are needed. Even humans can make mistakes sometimes, and in this case, it is important that the evaluation benchmark would allow the model to explore and return to previous status when a mistake is made and realized by the model. It is also interesting to explore how to automatically evaluate success rates for this task, \eg, by using LMMs~\citep{zhang2023gpt}. Another direction is to build GUI navigation datasets with different devices and diverse contents, \eg, personal computers and iPads.

\paragraph{Error correction.} A pretrained LMM may make mistakes due to data or algorithm bias. For example, if the agent fails to complete tasks in certain novel settings, how do we correct its errors to avoid mistakes in the future? Moreover, it would be interesting to study this in a continual learning setting, where the agent keeps interacting with new environments and receives new feedback continually.

\paragraph{Model distillation.} Using a large-scale model such as \gptname for GUI navigation is costly. In the future, it would be interesting to explore model distillation~\citep{polino2018model} for this task, to obtain a much smaller model with competitive navigation performance, which may achieve lower latency and higher efficiency.

% \paragraph{Different ways to mark.}
% \label{sec:grounding}
% OCR+iconNet
% Uniform grid with different grid sizes, SAM-different max number, Semantic-SAM with granularity, interactive user-added markers

% Quick discussion for completeness (although in practice, OCR+iconNet might make best sense)
\section{Conclusion}
% In this paper, we explore the task of phone screen navigation using \gptname, where the model is asked to take actions to complete an instruction on phone screens. and present a simple yet effective zero-shot baseline for this task. We conducted experiments on both iOS and Android systems, where \gptname presents a strong capability of environment understanding, reasoning and planning. Our work demonstrate the potential of building autonomous vision-language agents for device-control via pretrained vision-language models.  
% \zyang{future work: simulator and realistic multi-screen evaluation}
We have presented \modelname, a \gptname-based multimodal agent system designed for the GUI navigation task. The system is benchmarked on both our collected iOS dataset and a public Android navigation dataset, revealing \gptname's exceptional capabilities in understanding, reasoning, and planning over the screen environments. For future works, one promising direction is to establish a simulator-based benchmark that incorporates multi-step and episode-level automatic evaluations.
% \section*{Limitations}
% ACL 2023 requires all submissions to have a section titled ``Limitations'', for discussing the limitations of the paper as a complement to the discussion of strengths in the main text. This section should occur after the conclusion, but before the references. It will not count towards the page limit.
% The discussion of limitations is mandatory. Papers without a limitation section will be desk-rejected without review.

% While we are open to different types of limitations, just mentioning that a set of results have been shown for English only probably does not reflect what we expect. 
% Mentioning that the method works mostly for languages with limited morphology, like English, is a much better alternative.
% In addition, limitations such as low scalability to long text, the requirement of large GPU resources, or other things that inspire crucial further investigation are welcome.

% \section*{Ethics Statement}
% Scientific work published at ACL 2023 must comply with the ACL Ethics Policy.\footnote{\url{https://www.aclweb.org/portal/content/acl-code-ethics}} We encourage all authors to include an explicit ethics statement on the broader impact of the work, or other ethical considerations after the conclusion but before the references. The ethics statement will not count toward the page limit (8 pages for long, 4 pages for short papers).

% \section*{Acknowledgements}

\bibliography{anthology,custom}
\bibliographystyle{acl_natbib}

% \appendix

% \input{appendix/A-placeholder}

\end{document}